\documentclass[sigconf]{acmart}

\AtBeginDocument{%
  \providecommand\BibTeX{{%
    \normalfont B\kern-0.5em{\scshape i\kern-0.25em b}\kern-0.8em\TeX}}}

\copyrightyear{2023}
\acmYear{2023}
\setcopyright{rightsretained}
\acmConference[CIKM '23]{Proceedings of the 32nd ACM International Conference on Information and Knowledge Management}{October 21--25, 2023}{Birmingham, United Kingdom}
\acmBooktitle{Proceedings of the 32nd ACM International Conference on Information and Knowledge Management (CIKM '23), October 21--25, 2023, Birmingham, United Kingdom}
\acmDOI{10.1145/3583780.3615015}
\acmISBN{979-8-4007-0124-5/23/10}

\makeatletter
\gdef\@copyrightpermission{
  \begin{minipage}{0.3\columnwidth}
   \href{https://creativecommons.org/licenses/by/4.0/}{\includegraphics[width=0.90\textwidth]{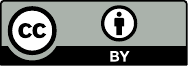}}
  \end{minipage}\hfill
  \begin{minipage}{0.7\columnwidth}
   \href{https://creativecommons.org/licenses/by/4.0/}{This work is licensed under a Creative Commons Attribution International 4.0 License.}
  \end{minipage}
  \vspace{5pt}
}
\makeatother

\usepackage{graphicx}
\usepackage{balance}
\usepackage{microtype}
\usepackage{booktabs}
\usepackage{array}
\usepackage{amsfonts}
\usepackage{amsmath}

\usepackage{amssymb}
\usepackage{multirow}
\usepackage{multicol}
\usepackage{listings}
\usepackage{subfigure}
\usepackage{graphicx}
\usepackage{enumitem}
\usepackage{hyperref}
\usepackage{xcolor,soul}

\newtheorem{defn}{Definition}
\newtheorem{obs}{Observation}

\begin{document}

\title{Prompt-and-Align: Prompt-Based Social Alignment for Few-Shot Fake News Detection}

\author{Jiaying Wu}
\affiliation{
  \institution{National University of Singapore}
  \country{}
  }
\email{jiayingwu@u.nus.edu}
\author{Shen Li}
\affiliation{
  \institution{National University of Singapore}
  \country{}
  }
\email{shen.li@u.nus.edu}
\author{Ailin Deng}
\affiliation{
  \institution{National University of Singapore}
  \country{}
  }
\email{ailin@u.nus.edu}
\author{Miao Xiong}
\affiliation{
  \institution{National University of Singapore}
  \country{}
  }
\email{miao.xiong@u.nus.edu}
\author{Bryan Hooi}
\affiliation{
  \institution{National University of Singapore}
  \country{}
  }
\email{bhooi@comp.nus.edu.sg}

\begin{abstract}

Despite considerable advances in automated fake news detection, due to the timely nature of news, it remains a critical open question how to effectively predict the veracity of news articles based on limited fact-checks. Existing approaches typically follow a \textit{``Train-from-Scratch''} paradigm, which is fundamentally bounded by the availability of large-scale annotated data. While expressive pre-trained language models (PLMs) have been adapted in a \textit{``Pre-Train-and-Fine-Tune''} manner, the inconsistency between pre-training and downstream objectives also requires costly task-specific supervision. In this paper, we propose \textit{``Prompt-and-Align''} ($\mathsf{P\&A}$), a novel prompt-based paradigm for few-shot fake news detection that jointly leverages the pre-trained knowledge in PLMs and the social context topology. Our approach mitigates label scarcity by wrapping the news article in a task-related textual prompt, which is then processed by the PLM to directly elicit task-specific knowledge. To supplement the PLM with social context without inducing additional training overheads, motivated by empirical observation on \emph{user veracity consistency} (i.e., social users tend to consume news of the same veracity type), we further construct a news proximity graph among news articles to capture the veracity-consistent signals in shared readerships, and align the prompting predictions along the graph edges in a confidence-informed manner. Extensive experiments on three real-world benchmarks demonstrate that $\mathsf{P\&A}$ sets new states-of-the-art for few-shot fake news detection performance by significant margins. \footnote{Data and code are available at: \url{https://github.com/jiayingwu19/Prompt-and-Align}.}

\end{abstract}



\maketitle

\section{Introduction}

The proliferation of fake news online poses an imperative concern for human cognition \cite{ecker2022psych,roberts21children} and social development \cite{pogue17how,sadiq22disaster}. Given the timeliness trait of news stories \cite{galtung65structure}, it is crucial that automated fake news detection applications enable accurate \textit{few-shot} veracity predictions based on limited related fact-checks. 

Nevertheless, the success of existing approaches is usually contingent on access to abundant fact-checked articles and auxiliary features, which is not guaranteed in practice. Regardless of whether utilizing the news content \cite{zhou2020safe,shu2019defend} or the social context graph \cite{cui2020deter,nguyen2020fang}, the majority of methods adopt a ``\textit{Train-from-Scratch}'' paradigm, where weight optimization is solely dependent on the supervised training data. Consequently, these methods typically require large-scale labeled news articles \cite{rashkin2017truth,shu2018fakenewsnet} and incorporate auxiliary information that are laborious to retrieve, including stance annotations \cite{nguyen2020fang}, user history posts \cite{dou2021user} and knowledge bases \cite{cui2020deter,dun2021kan,hu2021compare}. Under label scarcity, these methods suffer from generalization issues. To address this challenge, pre-trained language models (PLMs) \cite{devlin2019bert,liu2019roberta} have been adapted to the task following a \textit{``Pre-Train-and-Fine-Tune''} paradigm \cite{cruz2020localization,li2020connecting,pelrine2021surprising}, where a task-specific classification head is stacked upon a PLM. While methods under this paradigm benefit from the pre-trained syntactic and semantic knowledge, optimizing the auxiliary layers still requires abundant high-quality annotations, and fine-tuning a PLM alongside a randomly initialized task-specific architecture has been shown to distort the high-quality pre-trained features and impair model robustness \cite{hendrycks2020pretrained,kumar2022finetuning}. Under both paradigms, the real-world label scarcity of emerging news events creates a fundamental hurdle for weight optimization, resulting in substantial performance degradation.

\begin{figure}[t]
    \centering
    \includegraphics[width=\columnwidth]{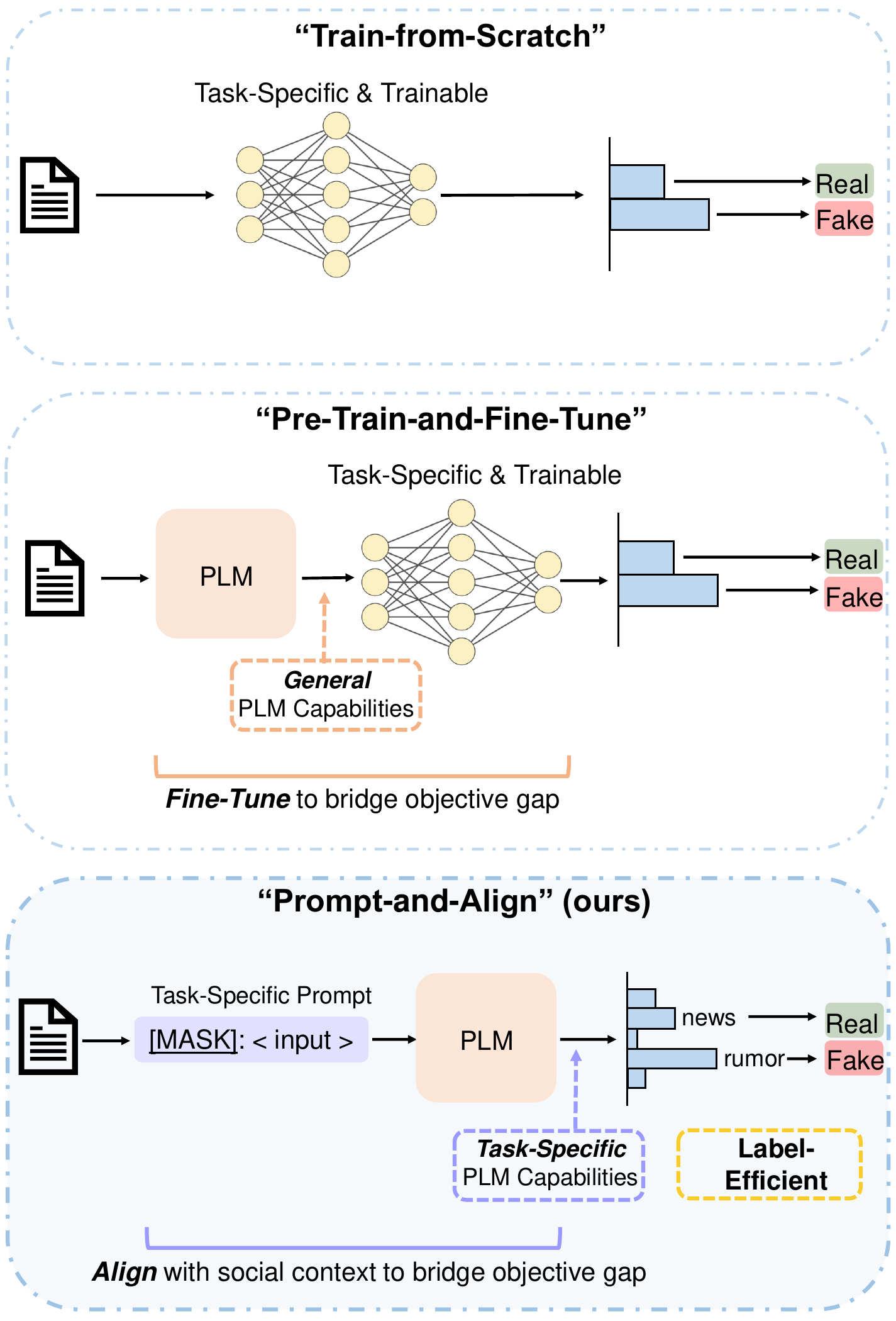}
    \caption{Comparison of our proposed \textit{``Prompt-and-Align''} ($\mathsf{P\&A}$) with existing paradigms.}
    \label{fig:paradigms}
\end{figure}

In this work, we develop \textit{``Prompt-and-Align''} ($\mathsf{P\&A}$), a novel prompt-based paradigm for few-shot fake news detection. Inspired by recent advances in prompt-based learning that exploit PLMs as powerful few-shot learners in various natural language processing tasks \cite{li2021prefix,petroni2019language,qin2021learning,schick2021exploiting}, we re-formulate fake news detection as a task-oriented text completion problem embedded in a natural language prompt. As illustrated in Figure \ref{fig:paradigms}, in contrast to existing ``Train-from-Scratch'' and ``Pre-Train-and-Fine-Tune'' paradigms that incorporate task-specific architectures, $\mathsf{P\&A}$ utilizes a textual prompt that encodes task-related knowledge. Prompting establishes semantic relevance between the task and the PLM \cite{liu2023pretrain}, elicits the latent ``built-in'' knowledge from PLMs for task-specific inference \cite{jiang2020know}, and thereby effectively alleviates the label scarcity bottleneck. 

On the basis of few-shot prompting, how can we further incorporate knowledge from the social context? An intuitive solution would be to combine prompting outputs with a Graph Neural Network (e.g., GCN \cite{kipf2017semi}) on the social graph. However, this does not address the problem, as the significant gap between the PLM pre-training objective and the downstream GNN classification objective can impair model performance. Across a sparsely labeled social graph, only a small portion of unlabeled nodes are involved in message passing, which fundamentally limits the effectiveness of the GNN.

To leverage informative structural patterns from the social context without inducing additional training overheads, we investigate the news consumption preference of social media users, and make a key observation on \textit{user veracity consistency}, where users are consistently attracted to news articles of a certain veracity type (i.e., real or fake). In other words, active social users connect multiple news articles with veracity-consistent signals via user engagements. Motivated by this empirical finding, we construct a news proximity graph to connect news articles with shared readership. To fully utilize the limited number of high-quality labels and high-confidence predictions, we first enhance the prompting predictions with ground-truth training data labels and soft labels generated via pseudo labeling \cite{lee2013pseudo}, and then align the independent predictions over the veracity-consistent graph edges, which further explicitly regularizes predictions of unlabeled samples to alleviate label scarcity.

To summarize, our contributions are three-fold:
\begin{itemize} [leftmargin=*]
\item \textbf{Empirical Finding}: We present a novel finding, on how \textit{user veracity consistency} leads to shared veracity between news articles with large common audiences.
\item \textbf{Method}: We propose \textit{``Prompt-and-Align"} ($\mathsf{P\&A}$), a novel paradigm for few-shot fake news detection that combines the benefits of both prompting and veracity-guided social alignment.
\item \textbf{Effectiveness}: Extensive experiments show that $\mathsf{P\&A}$ significantly enhances fake news detection accuracy by $8.93\%$, $11.01\%$, $6.36\%$ and $5.74\%$ on average under training set sizes $16$, $32$, $64$, and $128$, respectively. 
\end{itemize}

\section{Related Work}
Our work brings together two active lines of research.

\subsection{Fake News Detection on Social Media}
Deep learning models have shown impressive capacity for learning news representations. Existing methods can be generally categorized into two paradigms based on their training schemes: The dominant ``\textit{Train-from-Scratch}'' paradigm employs various neural architectures including Recurrent Neural Networks (RNNs) \cite{ruchansky2017csi,shu2019defend,wu2019different}, Convolutional Neural Networks (CNNs) \cite{wu2022probing,zhou2020safe} and Graph Neural Networks (GNNs) \cite{vaibhav2019sentence,monti2019fake,wu2023decor} to learn semantic and structural representations. To further enhance model prediction, existing methods incur high costs in retrieving various sources of auxiliary information including large-scale user response \cite{ruchansky2017csi}, entity descriptions from knowledge bases \cite{dun2021kan,hu2021compare}, open-domain evidence  \cite{sheng2021inter}, news producer description \cite{nguyen2020fang} and social media profiles \cite{dou2021user}.  Utilization of PLMs \cite{cruz2020localization,li2020connecting,pelrine2021surprising}, specifically fine-tuning them on annotated news samples, opened a new era of ``\textit{Pre-Train-and-Fine-Tune}'' with enhanced generalizability. However, existing methods typically incorporate task-specific architectures alongside a PLM, which require abundant labeled samples to optimize. 
Although initial progress has been made in meta-learning based multimodal fake news detection \cite{wang2021multi} involving news articles and images, the informative social context remains unconsidered. 
Hence, in this paper, we study the task of few-shot fake news detection on social media, which aims to effectively predict news veracity given a small number of annotated news samples and minimal social context (i.e., the numeric IDs of related social users).

\subsection{Prompt-Based Learning}

Recent years have realized PLMs \cite{brown2020language,devlin2019bert,liu2019roberta} as a strong impetus for advancing natural language understanding. Among PLM-based approaches, prompt-based learning has shown promising potential under few-shot scenarios. As per the investigations of \cite{petroni2019language,jiang2020know}, PLMs acquire abundant factual and commonsense knowledge during their pre-training stage, which can be elicited by re-formulating downstream tasks into text completion questions, either with manually pre-defined templates \cite{schick2021exploiting,tam2021improving} or with patterns learned on a small training set \cite{li2021prefix,qin2021learning}.
Guided by this knowledge, prompt-based methods have achieved impressive breakthroughs in a wide range of tasks including graph learning \cite{he2023explanations}, sentiment classification \cite{gao2021making}, natural language inference \cite{schick2021just},
relation extraction \cite{chen2022know}, and stance detection \cite{hardalov2022fewshot}.
Despite preliminary investigation into knowledge-based prompting for fake news classification \cite{jiang2022fake} (i.e., news documents augmented by auxiliary entity descriptions from a knowledge base as the PLM input), existing efforts overlook the informative social graph topology. On the related front of rumor detection, recent work leverages prompt learning for zero-shot transfer learning \cite{lin2022zero}. However, \cite{lin2022zero} focuses on learning language-agnostic contextual representations across different domains, which is inherently orthogonal to our contributions. In this work, we propose a novel \textit{``Prompt-and-Align''} ($\mathsf{P\&A}$) few-shot paradigm that addresses the generalization challenge by prompting the PLMs for task-related knowledge, and further alleviates the label scarcity issue via incorporating news readership patterns.

\section{Problem Definition}
\label{sec:problem}

Let $\mathcal{D}$ be a news dataset containing $N$ samples. $\mathcal{D}$ consists of the following components: \textit{\textbf{(1)}} a set of news articles $\mathcal{T}$ containing only textual information, with $|\mathcal{T}|=N$; \textit{\textbf{(2)}} a set of related social media users $\mathcal{U}$ who have reposted at least one article in $\mathcal{T}$; and \textit{\textbf{(3)}} a set of user engagements $\mathcal{R}$ in terms of user \textit{reposts}, where each engagement is formulated as $\{(u, T, r)|u\in \mathcal{U},T\in \mathcal{T}\}$ (i.e., user $u$ has engaged in spreading article $T$ via $r$ different reposts). Here, we assume minimal access to user attributes, restricting $\mathcal{U}$ to only a set of unique user IDs. For the few-shot setting, the article set $\mathcal{T}$ is split into a small labeled training set $\mathcal{T}_{L}$ of size $n$ and a large unlabeled test set $\mathcal{T}_{U}$. The articles in $\mathcal{T}_{L}$ are annotated with the corresponding one-hot labels $\mathcal{Y}_{L}=\{y_{1}, y_{2}, \ldots, y_{n}\}$, where $y_i\in\mathbb{R}^{C}$. In line with prior work \cite{cui2020deter,nguyen2020fang,shu2019defend}, we formulate fake news detection as a binary classification problem between the real and fake classes (i.e., $C=2$). 

Based on the above-mentioned notations, we formulate our task as follows:

\vspace{0.1in}
\fbox{
 \parbox{0.9\columnwidth}{
\underline{\textbf{\textit{Few-Shot Fake News Detection on Social Media:}}} 

\textbf{Input}: news dataset $\mathcal{D}=\{\mathcal{T},\mathcal{U},\mathcal{R}\}$, training labels $\mathcal{Y}_{L}$;

\textbf{Output}: predicted labels $\mathcal{Y}_{U}$.
} }

\begin{figure}[t]
    \centering
    \includegraphics[width=\columnwidth]{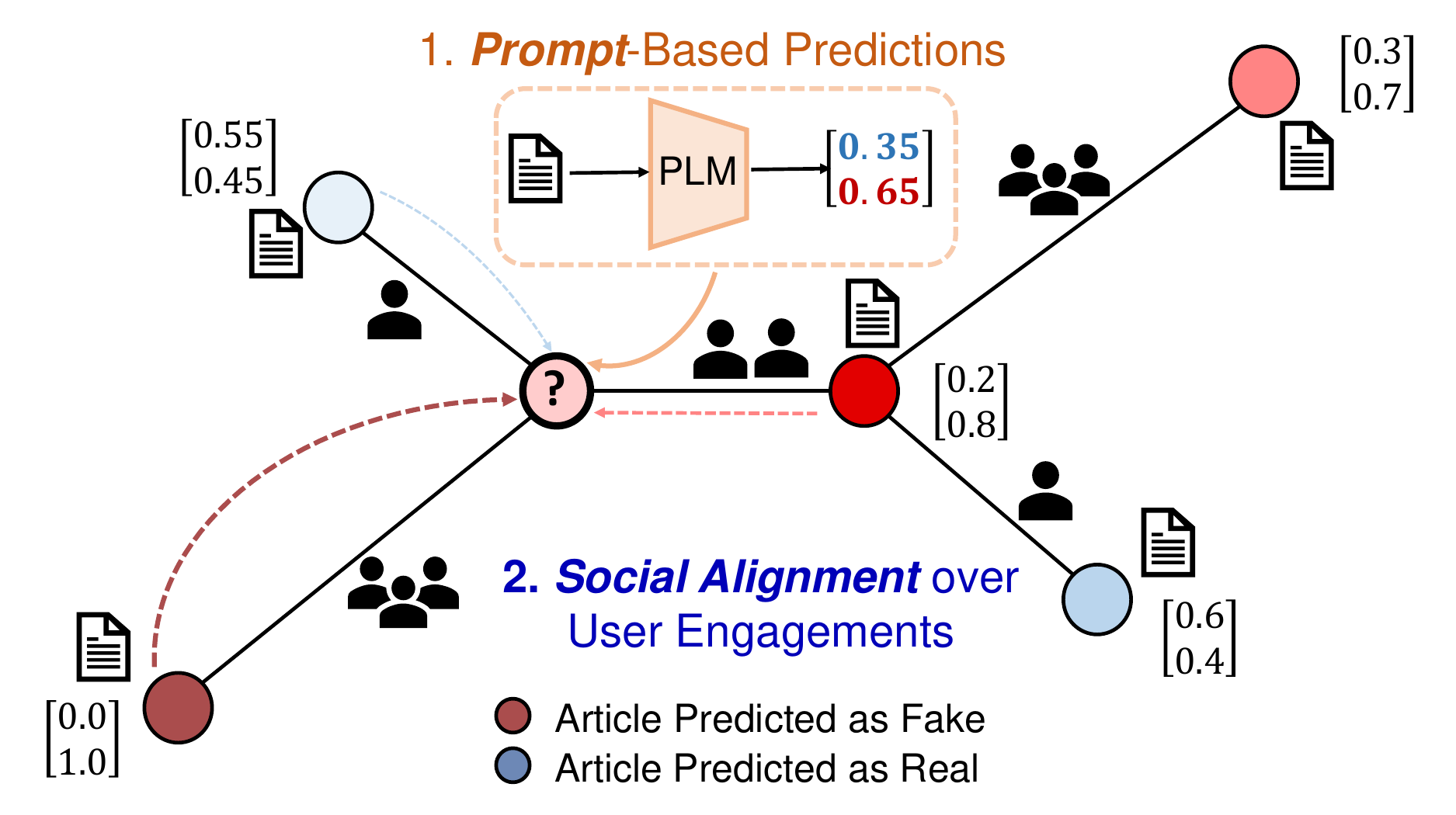}
    \caption{Overview of \textit{``Prompt-and-Align''} ($\mathsf{P\&A}$) paradigm. Veracity of the news article node with a question mark is determined by: (1) prompt-based predictions; and (2) social alignment over user engagements. News article nodes are connected by a news proximity graph (see Section \ref{sec:newsgraph}), and darker node colors denote higher confidence.}
    \label{fig:framework}
\end{figure}
\section{Proposed Approach}
\label{sec:approach}

In this section, we present our proposed ``Prompt-and-Align'' ($\mathsf{P\&A}$) paradigm for few-shot fake news detection, overviewed in Figure \ref{fig:framework}. $\mathsf{P\&A}$ consists of two major components, namely \textit{\textbf{(1)}} a \textbf{``Prompt''} component (Section \ref{sec:prompt}) that elicits task-specific knowledge from the PLM to predict news article veracity; and \textit{\textbf{(2)}} an \textbf{``Align''} component motivated by our empirical finding on user veracity consistency (Section \ref{sec:newsgraph}), which leverages the informative news readership patterns to enhance the prompting predictions via confidence-informed alignment over the social graph (Section \ref{sec:align}). 

\subsection{Prompting for Task-Specific Knowledge}\label{sec:prompt}

\begin{figure*}[t]
    \centering
    \subfigure[\textbf{PolitiFact}]{\includegraphics[width=0.27\textwidth]{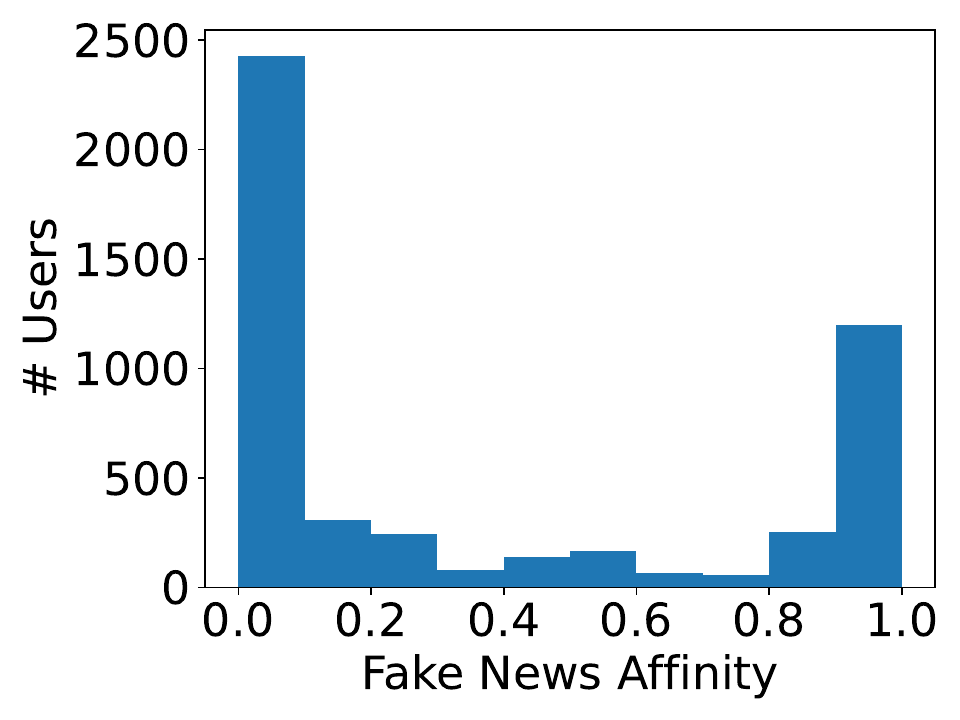}}
    \subfigure[\textbf{GossipCop}]{\includegraphics[width=0.27\textwidth]{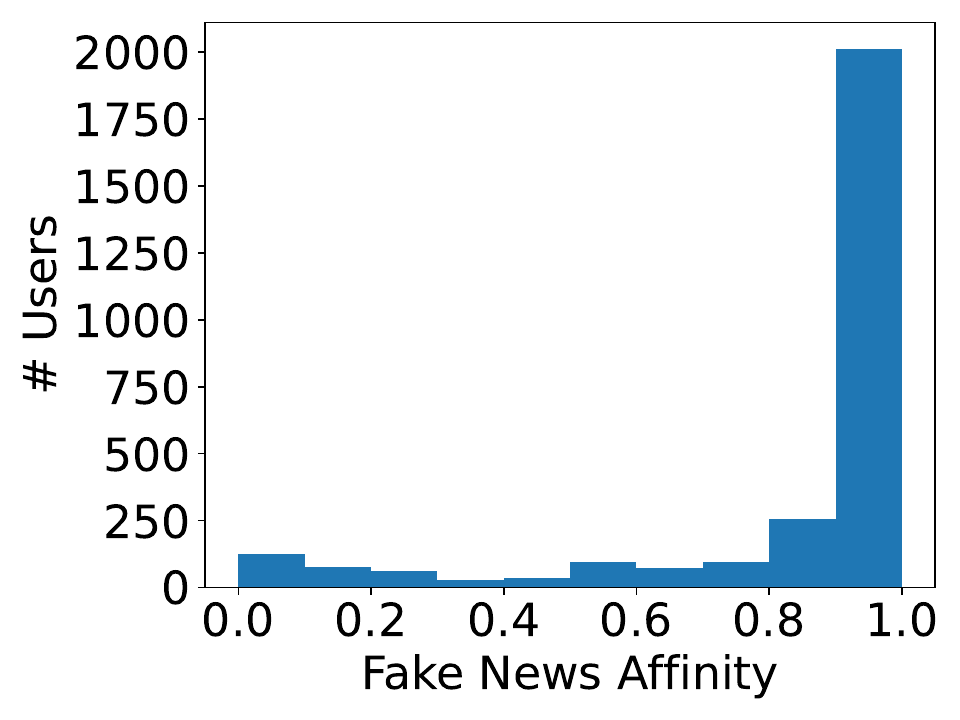}}
    \subfigure[\textbf{FANG}]{\includegraphics[width=0.27\textwidth]{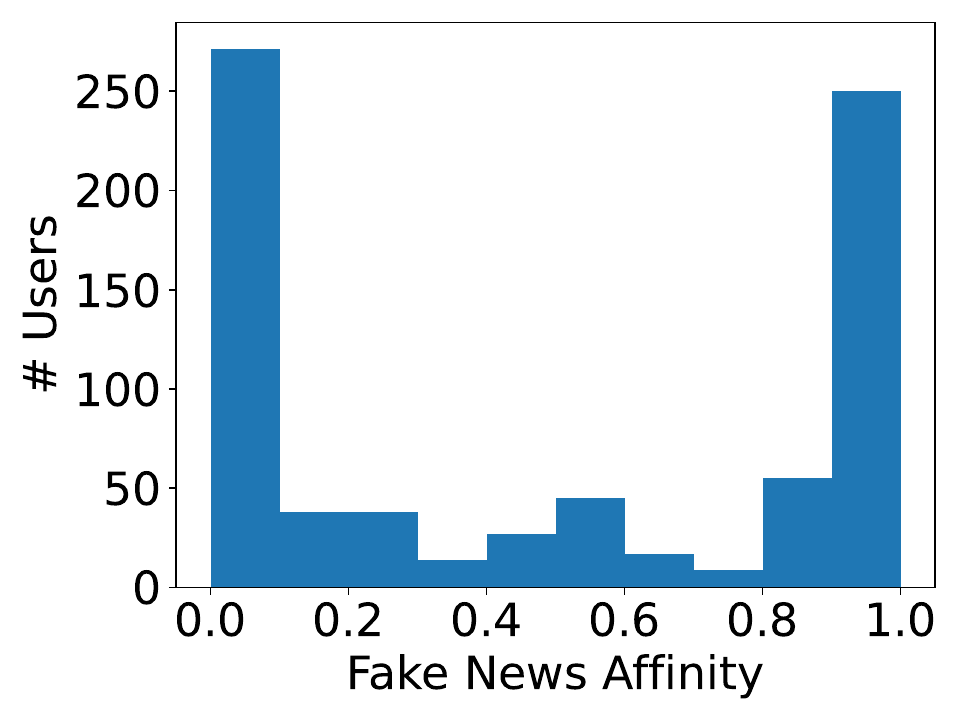}} 
    \caption{Social users exhibit a clear preference for either spreading fake news or real news. The fake news affinity score is formulated in Section \ref{sec:newsgraph}.
    }
    \label{fig:cred-hist}
\end{figure*}

The crucial factor in effectively connecting the PLM pre-training objective (i.e., masked language modeling) with the downstream fake news detection objective lies in identifying a shared template applicable to both tasks. To this end, we first re-formulate fake news detection as a \textit{masked token prediction} problem. 

Let $\mathcal{M}$ be a PLM with a vocabulary $\mathcal{V}$. Contrary to standard classification methods, which assign news articles to a label logit without any inherent meaning, prompting enables us to solve the fake news detection problem as follows: Given a news article $T_i$, we firstly construct a corresponding natural language prompt $\Tilde{T}_i$ containing a mask token:
\begin{equation}
    \Tilde{T}_i=\underline{\mathsf{[MASK]}}:T_i.
\label{eq:prompt}
\end{equation}

Taking $\Tilde{T}_i$ as input, $\mathcal{M}$ produces a score vector $S(\Tilde{T}_i)\in\mathbb{R}^{|\mathcal{V}|}$ over $\mathcal{V}$ at the masked position to complete the text. For token $v\in\mathcal{V}$, the corresponding score logit $S_v(\Tilde{T}_i)$ is computed as:
\begin{equation}
    S_v(\Tilde{T}_i)= \mathcal{M}(v|\Tilde{T}_i).
\label{eq:vocab_score}
\end{equation}
Here, the PLM is prompted to answer what is the most likely token that fits the $\mathsf{[MASK]}$ in $\Tilde{T}_i$. 

Given $S(\Tilde{T}_i)$, we define the answer space $\mathcal{A}=$\{``news'', ``rumor''\} as a small subset of the vocabulary containing permissible answers for our fake news detection prompt. Then, we map $\mathcal{A}$ to the binary labels following ``news''$\rightarrow0$ (real) and ``rumor''$\rightarrow1$ (fake), respectively.

To retrieve the conditional probability assigned to each answer token, we convert $S(\Tilde{T}_i)$ into a conditional probability distribution $P(\Tilde{T}_i)\in\mathbb{R}^{|\mathcal{V}|}$ over all vocabulary tokens via the softmax function. For answer token $y\in\mathcal{A}$, we extract its probability score $P_y(\Tilde{T}_i)$:
\begin{equation}
    P_y(\Tilde{T}_i)=\frac{\exp(S_y(\Tilde{T}_i))}{\sum_{v \in \mathcal{V}} \exp(S_v(\Tilde{T}_i))}.
\label{eq:token_score}
\end{equation}
Intuitively, our prompt-based approach incorporates task-specific semantic knowledge by prepending a task-related textual template to the input, and queries the PLM for whether it associates an article's text with news or rumor, by asking it to fill in the masked token with either ``news'' or ``rumor''. This helps directly elicit the model's task-specific knowledge, which significantly reduces the amount of training data it needs to learn effectively.

\noindent \textbf{Training objective.} The highest-scoring token in answer space $\mathcal{A}$ implies the PLM's prediction of news veracity. Hence, given the PLM $\mathcal{M}$ and our textual prompt $\Tilde{T}_i$, the goal is to maximize the probability score of the correct answer token $y_{+}$ and penalize the incorrect token $y_{-}$. For instance, real news have $y_{+}$ as ``news'' and $y_{-}$ as ``rumor''. The cross-entropy (CE) loss over answer space $\mathcal{A}$ is unsuitable here, as it only focuses on the two logits that correspond to the two answer tokens in $\mathcal{A}$. Hence, it fails to suppress the logits corresponding to non-answer tokens (e.g., task-irrelevant tokens such as ``coffee'' can be assigned a large logit for filling in $\mathsf{[MASK]}$), resulting in a suboptimal learned model. If we compute the CE loss over the entire vocabulary, this loss does not distinguish between $y_{-}$ and other non-answer tokens, and thus cannot specifically suppress $y_{-}$.  To address these issues, inspired by the decoupling label loss \cite{tam2021improving}, we adopt a loss function that simultaneously discourages non-answer tokens and keeps a specific focus on the answer token probabilities extracted from $P(\Tilde{T}_i)$ (see Eq. \ref{eq:token_score}). The loss can be formulated with the binary cross entropy (BCE) loss as:
\begin{equation}
    \mathcal{L}=\frac{1}{n}\sum_{i=1}^{n}\left(\mathsf{BCE}(P_{y_{+}}(\Tilde{T}_i), 1)+\mathsf{BCE}(P_{y_{-}}(\Tilde{T}_i), 0)\right),
\label{eq:loss}
\end{equation}
where $n$ is the number of annotated training samples.

\noindent \textbf{Base prediction acquisition.} To adapt the pre-trained weights to our fake news detection problem, we tune the parameters of $\mathcal{M}$ by minimizing the above loss function (Eq. \ref{eq:loss}) with the textual prompt, and retrieve a set of ``base prediction'' scores w.r.t. the $C$ class label tokens (mapped from the answer tokens) for all $N$ news articles in our dataset, denoted as $P_{\mathcal{A}}(\mathcal{T})\in \mathbb{R}^{N\times C}$. As the scores are extracted from a probabilistic distribution over all vocabulary tokens, we perform a softmax function to focus on the distribution over our task-specific answer space $\mathcal{A}$, formulated as a base prediction matrix $\mathbf{P}\in \mathbb{R}^{N\times C}$:
\begin{equation}
\mathbf{P}=\mathsf{Softmax}(P_{\mathcal{A}}(\mathcal{T})).
\label{eq:base-pred}
\end{equation}
At this point, we have $N$ independent prompting predictions obtained by probing the PLM with task-specific natural language prompts. Despite the rich semantic knowledge encapsulated in the PLM, the PLM does not contain social knowledge in terms of social graph topology. This motivates our innovations in further integrating auxiliary social context knowledge, specifically via distilling crowd wisdom from news readerships.

\subsection{Veracity-Consistent Readership Modeling}
\label{sec:newsgraph}

Social context, usually in the form of graphs, constitutes a distinctive property of the fake news detection problem \cite{shu2017fake}. In this subsection, we conduct preliminary analysis of real-world news engagements by social media users, with the aim of investigating the veracity-related social properties. Specifically, we explore the following question: \textbf{are there any connections between social users' news engagements and news veracity?}

To measure the social users' news consumption preference in terms of news veracity, we compute a \textit{``fake news affinity''} (FNA) score for user $u \in \mathcal{U}$, defined as:
\begin{defn}[Fake News Affinity] \label{defn:fna}
$$FNA(u)=\frac{\text{\# of fake news engagements by }u}{\text{\# of news engagements by }u}.$$
\end{defn}

\noindent \textbf{Empirical observation.} To capture patterns representative of user engagement preference, we set a user engagement threshold $t_u=5$, and focus solely on the active social users with at least $t_u$ engagements in spreading news articles. In Figure \ref{fig:cred-hist}, we visualize the FNA score distribution of active social users with a histogram, and make the following observation: 

\begin{obs} [User Veracity Consistency] \label{obs:user_veracity}
Active social users with numerous news engagements tend to have FNA scores either approaching $0$ (only engage in spreading real news) or $1$ (only engage in spreading fake news).
\end{obs}

Our observation echoes the confirmation bias theory \cite{nickerson1998confirm} for news consumption, where people tend to seek and retain information that reinforces their prior beliefs. Specifically, in the fake news detection task, confirmation bias can be manifested in terms of coordinated group behavior \cite{sharma19combating}, specifically with the aim to manipulate opinions on social platforms. For instance, social media has been shown to create opinion polarization towards political events \cite{kubin21role}, and users can coordinate boundaries to gain control over celebrity discussions on gossip sites \cite{mcnealy2019tea}. Consequently, user attention tends to become highly segregated on certain opinions, resulting in repeated engagements in news articles of a similar veracity type.

As an active social user engages in spreading numerous news pieces, the set of engaged news articles is connected with veracity-consistent signals. Therefore, if two news articles are closely connected with a large shared readership, the articles are highly likely to have the same veracity label.

\noindent\textbf{News Proximity Graph.} Guided by our empirical observation, we construct a news proximity graph $\mathcal{G}$ that encodes the shared readerships between new article nodes $\mathcal{T}$. Let $\mathbf{A}_{\mathcal{T}}\in\mathbb{R}^{N\times N}$ be the adjacency matrix of $\mathcal{G}$ that quantifies shared readerships, which we will next derive from the set $\mathcal{R}$ of user repost records. 

To focus on the active social users, we set a user engagement threshold $t_u$ to remove the users with fewer than $t_u$ news reposts. From $\mathcal{R}$, we construct a user engagement matrix $\mathbf{B}\in\mathbb{R}^{|\mathcal{U}'|\times N}$, where $\mathcal{U}' \subseteq \mathcal{U}$ filtered via user engagement threshold $t_u$ denotes the active user set, and element $\mathbf{B}^{km}$ denotes the engagement intensity of user $u_k\in\mathcal{U}'$ towards news article $T_m\in\mathcal{T}$. The value of $\mathbf{B}^{km}$ is obtained from $r_{km}$ in the corresponding entry $(u_k,T_m,r_{km})\in \mathcal{R}$ in the user engagement set.

On the basis of user engagement matrix $\mathbf{B}$, we formulate the news proximity matrix $\mathbf{A}_{n}\in\mathbb{R}^{N\times N}$ as $\mathbf{B}^\top \mathbf{B}$. Then, we conduct normalization on $\mathbf{A}_{n}$ to derive:
\begin{equation}
    \mathbf{A}_{\mathcal{T}} = \mathbf{D}_{n}^{-\frac{1}{2}} \mathbf{A}_{n} \mathbf{D}_{n}^{-\frac{1}{2}}.
\end{equation}

\subsection{Confidence-Informed Social Alignment}
\label{sec:align}

To inject social context into the prompting predictions, motivated by our empirical finding on user veracity consistency (Section \ref{sec:newsgraph}), we construct a news proximity graph $\mathcal{G}$ to connect news article nodes with veracity-consistent signals from shared readerships. However, training a task-oriented GNN (e.g., GCN \cite{kipf2017semi}) on $\mathcal{G}$ is inapplicable under label sparsity, as the involvement of unlabeled nodes is heavily limited during message passing, resulting in poor model performance. To fully exploit label knowledge and social graph structure without inducing additional trainable modules, we propose a social alignment component that combines the base predictions in a confidence-informed manner. In addition to prompting, our graph-based alignment further bridges the gap between general PLM pre-training and the downstream fake news detection task, which enhances the effectiveness of our approach.

\subsubsection{Label Knowledge Acquisition.}

We first discuss how our alignment component acquires knowledge from the relatively ``cheap'' base predictions $\mathbf{P}$ derived from task-specific prompting. The key idea is that we leverage the ground truth training labels and a small subset of unlabeled samples (i.e., news articles) with high prediction confidence. To this end, we transform $\mathbf{P}$ into an enhanced confidence matrix $\mathbf{H}\in \mathbb{R}^{N\times C}$. Specifically, $\mathbf{H}$ consists of $\mathbf{H}_L \in \mathbb{R}^{n\times C}$ and $\mathbf{H}_U \in \mathbb{R}^{(N-n)\times C}$, which respectively refer to the enhanced confidence scores of labeled and unlabeled samples. $\mathbf{H}$ is computed via the following two steps.

\noindent\textbf{Incorporation of ground-truth knowledge.} To fully exploit the high-quality annotated training samples $\mathcal{T}_{L}$, we stack the one-hot training data labels $\mathcal{Y}_{L}$ into a ground truth matrix $\mathbf{Y}_{L}\in \mathbb{R}^{n\times C}$, and utilize the ground-truth class probabilities to replace the base predictions $\mathbf{P}_{L}\in \mathbb{R}^{n\times C}$ pertaining to the training data. The corresponding rows in $\mathbf{H}$ is formulated as:
\begin{equation}
    \mathbf{H}_{L} = \mathbf{Y}_{L}.
\label{eq:train-aug}
\end{equation}

\noindent \textbf{Thresholded pseudo labeling.} The prompting module of $\mathsf{P\&A}$ enables access to abundant soft labels assigned by the PLM, presented in the form of base predictions $\mathbf{P}_{U}$ for unlabeled data $\mathcal{T}_{U}$. To fully utilize the soft labels, we devise a thresholded variant of the pseudo labeling technique \cite{lee2013pseudo}, denoted as $\mathsf{ThresholdedPL}(\cdot)$. Given $\mathbf{P}_{U}$, we select a set of samples with high class probability scores equal to or above the $t_p$-th percentile of all predicted confidence scores in $\mathbf{P}_{U}$. Following this, we assign these high-confidence samples with one-hot labels w.r.t. the class with maximum predicted probability for each sample, while the remaining low-confidence samples are kept unchanged. We augment $\mathbf{P}_{U}$ with the resultant pseudo labels to obtain enhanced confidence scores $\mathbf{H}_{U}$:
\begin{equation}
    \mathbf{H}_{U} = \mathsf{ThresholdedPL}(\mathbf{P}_{U}).
\label{eq:thresholded-pl}
\end{equation}

\subsubsection{Veracity-Guided Prediction Alignment} Knowledge-infused predictions in $\mathbf{H}$ facilitate effective utilization of the limited high-quality ground truth labels and predictions. Hence, given $\mathbf{H}$ and the veracity-consistent edges of news proximity graph $\mathcal{G}$, we conduct veracity-guided social alignment to combine the independent predictions in a confidence-informed manner. 

Specifically, we encourage label smoothness over the graph structure by aligning $\mathbf{H}$ via propagation over the $k$-hop neighborhood of each news article node. From the propagated scores, we obtain the final aligned predictions $\mathbf{\hat{Y}}$:
\begin{equation}
    \mathbf{\hat{Y}}=A_{\mathcal{T}}^{k}\mathbf{H},
\label{eq:label-align}
\end{equation}
among which the predicted class label of news article $T_m\in\mathcal{T}_{U}$ is assigned by
\begin{equation}
    \hat{y}_m=\mathsf{argmax}_{j \in\{1, \ldots, C\}} \hat{\mathbf{Y}}_{mj}.
\end{equation}

Here, the obtained $\hat{y}_m$ aggregates social knowledge (specifically, veracity signals) through our proposed social alignment component, which thereby refines the base predictions derived from prompting.

\begin{table}[t]
\caption{Dataset statistics.}
 \begin{tabular}{lccc} \toprule
 \textbf{Dataset} &  \textbf{PolitiFact} & \textbf{GossipCop} & \textbf{FANG} \\ 
 \toprule
 \# News Articles & 482 & 1,736 & 664 \\
 \# Real News & 241 & 868 & 332 \\
 \# Fake News & 241 & 868 & 332 \\  
 \# User Engagements & 236,950 & 163,532 & 50,549 \\ 
 \# Distinct Users & 148,023 & 79,474 & 37,165 \\
 \bottomrule
\end{tabular} 
 \label{tab:ds-stats}
\end{table}

\begin{table*}[ht]
\caption{Performance comparison between $\mathsf{P\&A}$ and baselines in terms of fake news detection accuracy (\%) at varying training set sizes ($n\in \{16,32,64,128\}$). Bold (underline) indicates the best overall (baseline) performance. $^{*}$ denotes that  improvements of $\mathsf{P\&A}$ over the most competitive baselines are significant at $p<0.01$ level under the Wilcoxon signed-rank test \cite{wilcoxon1945rank}.}
\centering 
 \begin{tabular}{lcccccccccccccc} \toprule
 \multirow{2}{*}{\textbf{Method}} & \multicolumn{4}{c}{\textbf{PolitiFact}} && \multicolumn{4}{c}{\textbf{GossipCop}} && \multicolumn{4}{c}{\textbf{FANG}}\\
 \cmidrule{2-5} \cmidrule{7-10} \cmidrule{12-15}
 & 16 & 32 & 64 & 128 &&  16 & 32 & 64 & 128 && 16 & 32 & 64 & 128 \\ 
 \toprule
 dEFEND\textbackslash c \cite{shu2019defend}  & 51.78 & 54.62 & 61.06 & 66.34 &&  50.43 & 50.35 & 51.48 & 52.66 && 50.19 & 50.85 & 50.91 & 54.60 \\
 SAFE\textbackslash v \cite{zhou2020safe} & 51.71 & 54.52 & 61.91 & 68.36 &&  51.35 & 51.20 & 53.39 & 57.40 && 51.90 & 53.01 & 55.53 & 58.35 \\
 SentGCN \cite{vaibhav2019sentence}  & 56.34 & 51.95 & 56.81 & 56.23 &&  49.57 & 49.33 & 50.04 & 53.87 && 51.35 & 50.60 & 52.58 & 54.45 \\
 SentGAT \cite{vaibhav2019sentence} & 56.09 & 52.99 & 55.64 & 58.36 &&  49.54 & 50.11 & 50.52 & 54.65 && 51.04 & 51.59 & 54.09 & 56.34 \\
 FANG \cite{nguyen2020fang}  & 57.87 & 60.45 & 68.72 & 77.56 &&  51.73 & 54.10 & \underline{63.93} & 69.65 && 53.32 & 56.22 & 58.53 & 60.49 \\
 GCNFN \cite{schick2021exploiting} & 55.37 & 61.86 & 70.06 & \underline{84.46} &&  52.18 & 53.85 & 62.29 & \underline{71.89} && 52.43 & 56.08 & 60.12 & \underline{61.85} \\
 \midrule
 BERT-FT \cite{devlin2019bert} & 61.27 & 67.48 & 73.52 & 77.43 &&  52.45 & 52.74 & 55.02 & 59.28 && 53.28 & 54.71 & 57.02 & 58.41 \\
 RoBERTa-FT \cite{liu2019roberta} & 54.80 & 61.16 & 78.95 & 81.44 &&  52.54 & 54.17 & 54.09 & 61.35 && 51.47 & 54.56 & 56.92 & 60.92 \\
 PET \cite{schick2021exploiting} & 64.24 & 68.09 & 79.44 & 80.49 &&  53.69 & \underline{55.11} & 59.78 & 62.97 && 55.76 & \underline{56.46} & 59.13 & 59.79 \\
 KPT \cite{hu22knowledgeable} & \underline{68.27} & \underline{70.21} & \underline{80.40} & 83.16 &&  \underline{53.95} & 54.70 & 60.12 & 62.19 && \underline{56.87} & 55.79 & \underline{60.36} & 61.38 \\
 \midrule
 $\mathsf{P\&A}$ (ours) & \textbf{85.30$^{*}$} & \textbf{83.73$^{*}$} & \textbf{86.79$^{*}$} & \textbf{89.45$^{*}$} &&  \textbf{61.15}$^{*}$ & \textbf{70.59$^{*}$} & \textbf{74.62$^{*}$} & \textbf{81.60$^{*}$} && \textbf{59.42$^{*}$} & \textbf{60.48$^{*}$} & \textbf{62.37$^{*}$} & \textbf{64.37$^{*}$} \\

 \bottomrule
\end{tabular} 
\label{tab:performance}
\end{table*}

\section{Experiments}

In this section, we empirically evaluate our \textit{``Prompt-and-Align''} ($\mathsf{P\&A}$) paradigm to investigate the following five research questions:

\begin{itemize} [leftmargin=*]
    \item \textbf{Few-Shot Performance} (Section \ref{sec:fnd_performance}): How effective is $\mathsf{P\&A}$ in few-shot fake news detection?
    \item \textbf{Ablation Study} (Section \ref{sec:ablation}): How do news content and social graph structure contribute to the performance of $\mathsf{P\&A}$?
    \item \textbf{Parameter Sensitivity Analysis} (Section \ref{sec:param_sensitivity}): How does $\mathsf{P\&A}$ perform under different alignment steps, pseudo labeling percentiles, and user engagement thresholds?    \item \textbf{Impact of Prompt Design} (Section \ref{sec:prompt_templates}): How effective is $\mathsf{P\&A}$ across different textual prompts?
    \item \textbf{Case Study} (Section \ref{sec:case_study}): How can we interpret the predictions of $\mathsf{P\&A}$ via its intermediate results?
\end{itemize}

\subsection{Experimental Setup} 
\subsubsection{Datasets} 
We conduct evaluation on three real-world benchmark datasets commonly adopted by existing work, namely the FakeNewsNet \cite{shu2018fakenewsnet} public benchmark consisting of the \textbf{PolitiFact} and \textbf{GossipCop} datasets, and the \textbf{FANG} \cite{nguyen2020fang} dataset. All datasets contain news articles collected from leading fact-checking websites and the related social user engagements (i.e., IDs of repost users) retrieved from Twitter. The dataset statistics are shown in Table \ref{tab:ds-stats}. 

We split the data into training and test sets by randomly sampling $n$ news items as the training data (with $n\in \{16, 32, 64, 128\}$), among which the ratio of (fake news) / (real news) is set as $1:1$.

\subsubsection{Baselines} 
\label{sec:baselines}

We benchmark $\mathsf{P\&A}$ against ten representative baseline methods that adopt the following paradigms:

\textbf{``Train-from-Scratch'' methods} devise task-specific neural architectures for fake news detection. \textbf{dEFEND}\textbackslash c is a variant of the hierarchical attention framework dEFEND \cite{shu2019defend} without user comments. \textbf{SAFE}\textbackslash v is a variant of SAFE \cite{zhou2020safe} without the visual component, which adopts a TextCNN \cite{kim14convolutional} based module to encode the news article texts. \textbf{SentGCN} and \textbf{SentGAT} \cite{vaibhav2019sentence} are graph-based approaches that respectively employ the Graph Convolutional Network (GCN) \cite{kipf2017semi} and the Graph Attention Network (GAT) \cite{velickovic2018graph} to capture indicative sentence interaction patterns. \textbf{GCNFN} \cite{monti2019fake} utilizes deep geometric learning to model news dissemination patterns along with textual node embedding features. \textbf{FANG} \cite{nguyen2020fang} constructs a heterogeneous social graph with news articles, sources and social users, and adopts a GraphSAGE \cite{hamilton2017inductive} based framework to detect fake news. For a fair comparison, social context based methods are implemented with the components for encoding the news content, user-news relations and social user identities.

\textbf{PLM-based methods} leverage the rich pre-trained knowledge in PLMs to mitigate label scarcity. Among this category, \textit{``Pre-Train-and-Fine-Tune''} methods \textbf{BERT-FT} and \textbf{RoBERTa-FT} respectively combine  BERT \cite{devlin2019bert} and RoBERTa \cite{liu2019roberta} models with a task-specific MLP to predict news veracity. \textit{Prompt-tuning} methods include \textbf{PET} \cite{schick2021exploiting}, which provides task descriptions to PLMs for supervised training via task-related cloze questions and verbalizers; and \textbf{KPT} \cite{hu22knowledgeable}, which expands the label word space with class-related tokens of varied granularities and perspectives. For a fair comparison, we do not implement self-training and PLM ensemble for PET. We utilize the base versions of BERT and RoBERTa for fine-tuning, and adopt BERT-base as the backbone of prompt-tuning baselines, consistent with the setup of our proposed approach.

As $\mathsf{P\&A}$ focuses on news content and social context, we do not compare $\mathsf{P\&A}$ with knowledge-based approaches \cite{jiang2022fake,dun2021kan,hu2021compare} that incorporate entity information from external knowledge bases, which are orthogonal to our contributions.

\subsubsection{Implementation Details}
We implement $\mathsf{P\&A}$ and its variants based on PyTorch 1.8.0 with CUDA 11.1. We utilize pretrained BERT-base weights from HuggingFace Transformers 4.13.0 \cite{wolf20transformers}. The maximum sequence length, batch size, and learning rate are set to $512$, $16$, and $5\times10^{-5}$ respectively, consistent with \cite{devlin2019bert}. We fine-tune the models for $3$ epochs for $n\in \{16, 32\}$, and $5$ epochs for $n\in \{64, 128\}$. In the ``Align''  module, we set the pseudo labeling threshold $t_p$ at the $95$-th percentile among all test data predictions, and set the user engagement threshold $t_u$ to $5$. The number of alignment steps $k$ is set to $2$. For the baseline methods, we follow the architectures and hyperparameter values suggested by their respective authors. In all experiments, we report the average test accuracy (\%) across 20 different runs of each method.

\begin{figure*}[t]
    \centering
    \subfigure[\textbf{PolitiFact}]{\includegraphics[width=0.32\textwidth]{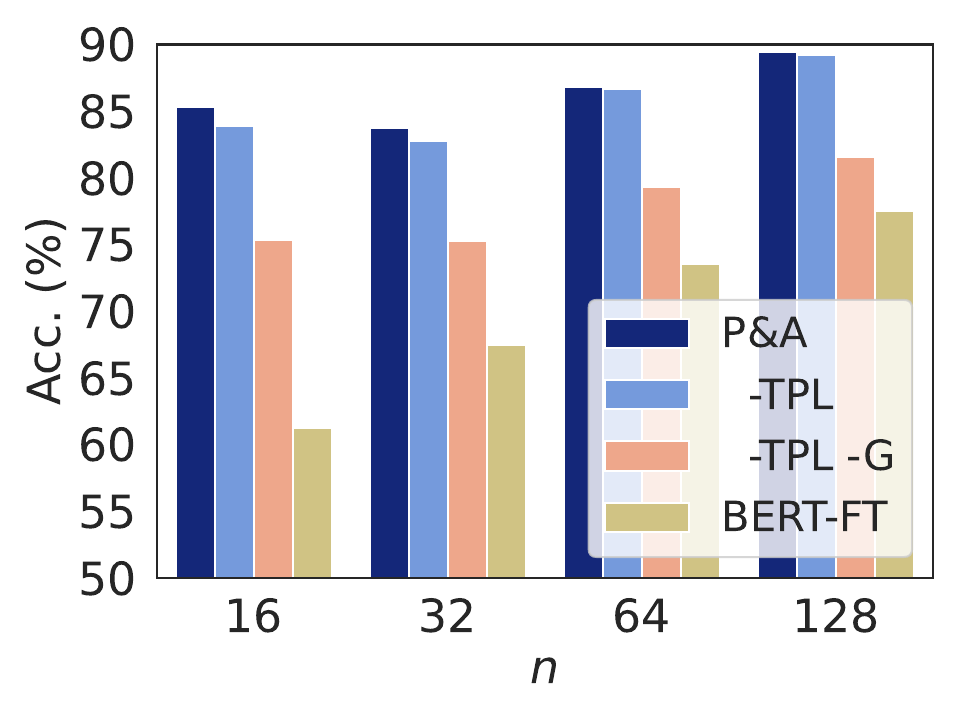}}
    \subfigure[\textbf{GossipCop}]{\includegraphics[width=0.32\textwidth]{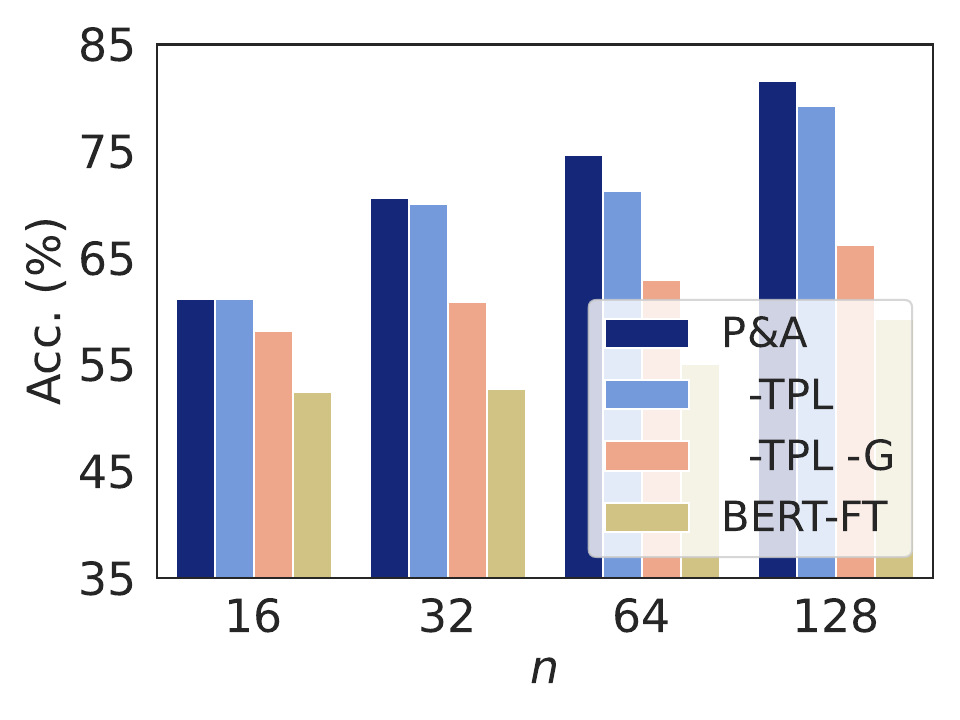}}
    \subfigure[\textbf{FANG}]{\includegraphics[width=0.32\textwidth]{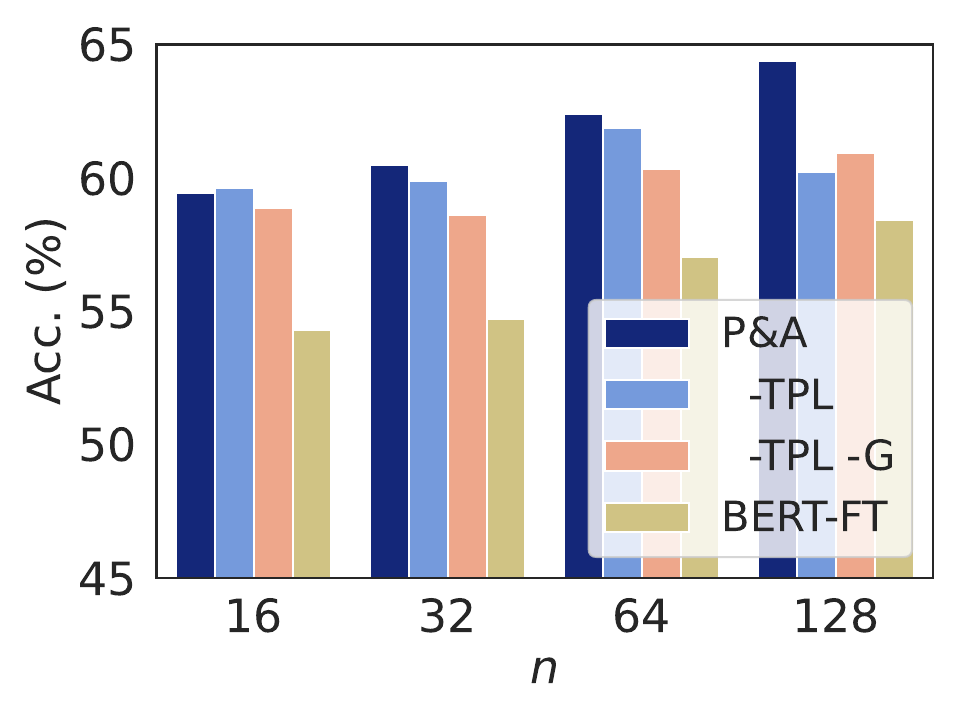}} 
    \caption{Ablation study of $\mathsf{P\&A}$ at training set size $n$. 
    ``-TPL'' removes thresholded pseudo labeling (see Eq. \ref{eq:thresholded-pl}), ``-TPL -G'' only implements the ``Prompt'' component, and ``BERT-FT'' refers to the BERT model fine-tuned with task-specific layers.
    }
    \label{fig:ablation}
\end{figure*}

\subsection{Few-Shot Detection Performance}
\label{sec:fnd_performance}

Table \ref{tab:performance} compares the performance of $\mathsf{P\&A}$ with competitive baselines. We can observe that: 
\textit{\textbf{(1)}} Among ``Train-from-Scratch'' methods, social graph based methods (FANG and GCNFN) consistently outperform news content based methods, which indicates the importance of incorporating social context topology. 
\textit{\textbf{(2)}} PLM-based methods outperform ``Train-from-Scratch'' methods in numerous cases, showing that the informative pre-trained features in PLMs can help to address the label scarcity issue.
\textit{\textbf{(3)}} Among PLM-based methods, prompting methods (PET and KPT) consistently outperform the standard fine-tuned PLMs, which justifies our analysis that prompting directly elicits task-specific knowledge from PLMs.
\textit{\textbf{(4)}} $\mathsf{P\&A}$ substantially outperforms the most competitive baseline. Given $n$ training samples ($n \in \{16,32,64,128\}$), $\mathsf{P\&A}$ respectively enhances few-shot fake news detection accuracy by $8.93\%$, $11.01\%$, $6.36\%$ and $5.74\%$ on average. The statistically significant performance gains validate the effectiveness of eliciting task-specific knowledge with prompts and performing confidence-informed prediction alignment across the news proximity graph.

\subsection{Ablation Study}
\label{sec:ablation}

To investigate the contribution of $\mathsf{P\&A}$ components, we ablate two important components of our approach (thresholded pseudo labeling and social graph alignment) via the following variants:
\begin{itemize} [leftmargin=*]
    \item $\mathsf{P\&A (-TPL)}$, which removes the thresholded pseudo labeling mechanism on high-confidence samples (see Eq. \ref{eq:thresholded-pl}).
    \item $\mathsf{P\&A (-TPL -G)}$, which eliminates the social context (i.e., our news proximity graph) and reduces to the ``Prompt'' component.
\end{itemize}

Note that when we only remove the graph-based social alignment steps (i.e., $\mathsf{(-G)}$), the results are the same as $\mathsf{(-TPL -G)}$, as thresholded pseudo labeling does not invert any predicted labels. 

Figure \ref{fig:ablation} shows the few-shot performance of different $\mathsf{P\&A}$ variants on three datasets. We find that \textit{\textbf{(1)}} Social alignment significantly contributes to overall performance, which indicates that aligning confidence-enhanced prompting predictions across our news proximity graph effectively utilizes veracity-consistent signals in the social context. \textit{\textbf{(2)}} Removing the thresholded pseudo labeling mechanism leads to inferior performance than $\mathsf{P\&A}$, which is expected as removing pseudo labels exacerbates label sparsity on the social graph. \textit{\textbf{(3)}} $\mathsf{P\&A (-TPL -G)}$ (i.e., our prompting module) consistently outperforms BERT-FT, which is consistent with our analysis that tuning the PLM via task-specific textual prompts helps to directly elicit task-specific knowledge and alleviate label scarcity.

\begin{figure*}[t]
    \centering
    \subfigure[\textbf{$n=16$}]{\includegraphics[width=0.24\textwidth]{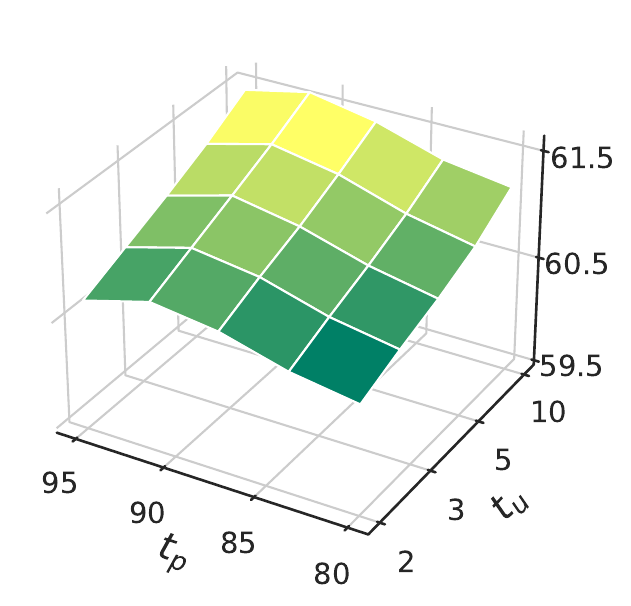}}
    \subfigure[\textbf{$n=32$}]{\includegraphics[width=0.24\textwidth]{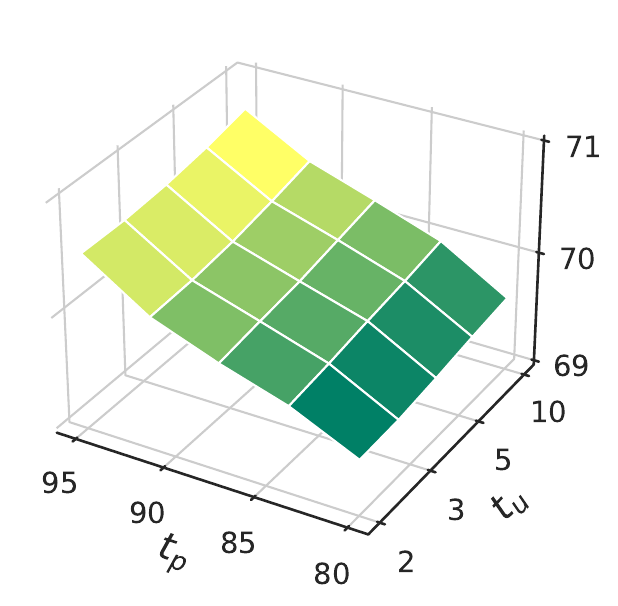}}
    \subfigure[\textbf{$n=64$}]{\includegraphics[width=0.24\textwidth]{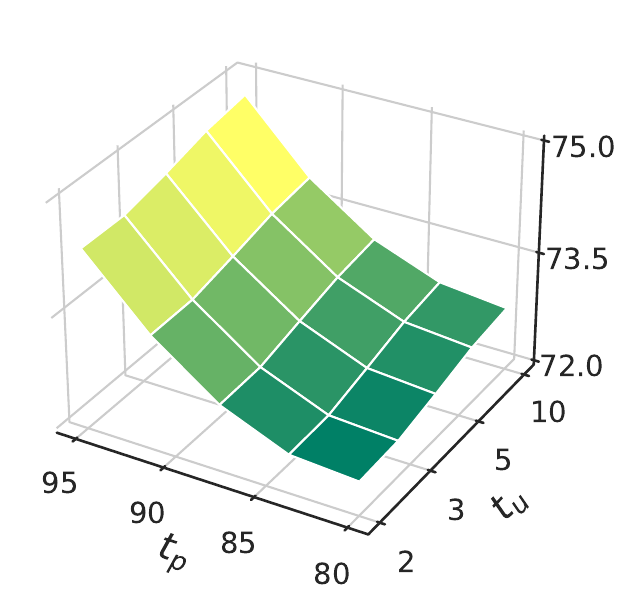}}
    \subfigure[\textbf{$n=128$}]{\includegraphics[width=0.24\textwidth]{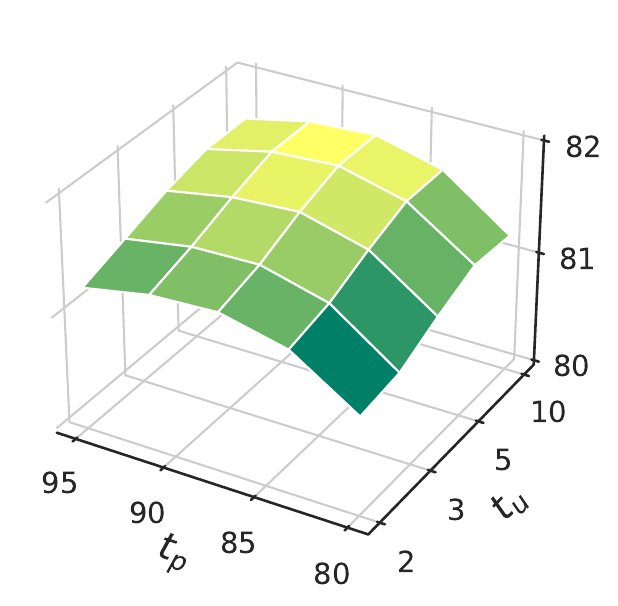}}
    \caption{Parameter sensitivity analysis of $\mathsf{P\&A}$ on GossipCop under different combinations of $t_p$ (pseudo labeling percentile) and $t_u$ (user engagement threshold) values. Lighter color represents higher accuracy (z-axis, \%).} 
    \label{fig:robustness}
\end{figure*}

\subsection{Parameter Sensitivity Analysis}
\label{sec:param_sensitivity}
We explore the sensitivity of three important hyperparameters in $\mathsf{P\&A}$: pseudo labeling percentile $t_p$ for leveraging high-confidence predictions, user engagement threshold $t_u$ for filtering inactive users, and alignment steps $k$ over the news proximity graph.

\subsubsection{Effects of Threshold Values}
To evaluate the impact of threshold values $t_p$ and $t_u$, we investigate the performance of $\mathsf{P\&A}$ with $t_p \in \{80, 85, 90, 95\}$ and $t_u \in \{2,3,5,10\}$, and present the results in Figure \ref{fig:robustness}. We observe that: \textit{\textbf{(1)}} The performance of $\mathsf{P\&A}$ remains relevantly smooth across different variations of threshold values, with all combinations significantly outperforming the best baselines. Our default threshold values of $t_p=95$ and $t_u=5$ consistently give satisfactory performance, which indicates that our settings for the social alignment stage are practical for few-shot fake news detection on social media. \textit{\textbf{(2)}} When we gradually lower the pseudo labeling threshold $t_p$, the performance of $\mathsf{P\&A}$ slightly degrades. This is because $t_p$ is formulated as a percentile, with which $\mathsf{P\&A}$ automatically determines the numeric threshold based on the distribution of all prediction scores. Consequently, $\mathsf{P\&A}$ assigns one-hot labels to more samples as $t_p$ decreases, whose inherent uncertainty leads to error accumulation.

\subsubsection{Effects of Graph Neighborhood Sizes} As validated by our ablation study in Section \ref{sec:ablation}, the veracity-consistent social context (i.e., news proximity) effectively enhances the prompting predictions. To investigate the structural news readership patterns embedded in the news proximity graph, we evaluate $\mathsf{P\&A}$ performance over varying numbers of alignment steps (denoted as $k$ in Eq. \ref{eq:label-align}), with $k \in\{1,2,3,4\}$. 

As shown in Table \ref{tab:k-hop}, we find that: \textit{\textbf{(1)}} The performance of $\mathsf{P\&A}$ remains relevantly smooth across different variations of $k$, with all combinations significantly outperforming the best baselines. Our proposed $\mathsf{P\&A}$ adopts $k=2$ alignment steps across the news proximity graph, which consistently yields satisfactory results. \textit{\textbf{(2)}} In addition to 1-hop neighborhood (i.e., between news articles with direct overlaps of readerships), aggregating information from each article's 2-hop and 3-hop neighborhoods on the news proximity graph can boost the performance of $\mathsf{P\&A}$, which indicates the effectiveness of higher-order social context modeling. \textit{\textbf{(3)}} Compared with incorporating 2-hop and 3-hop neighbors, incorporating 4-hop neighborhoods leads to slight performance degradation. Indeed, large neighborhood sizes might not be helpful for fake news detection, as they contain intermingled veracity signals from the ``fake'' and ``real'' classes, which obstructs prediction alignment.

\begin{table}[t]
\caption{Accuracy (\%) of $\mathsf{P\&A}$ over different numbers of alignment steps on the news proximity graph. 
}
\centering 
 \begin{tabular}{lccccc} \toprule
 Dataset & $k$ & $n=16$ & $n=32$ & $n=64$ & $n=128$\\
 \toprule
 \multirow{4}{*}{PolitiFact} & 
 1 & 83.15 & 82.25 & 87.03 & 90.11  \\ &
 2 & 85.30 & 83.73 & 86.79 & 89.45  \\ &
 3 & 85.44 & 83.18 & 86.38 & 88.81 \\& 
 4 & 85.34 & 82.44 & 85.68 & 88.32  \\ 
 \midrule
 \multirow{4}{*}{FANG} & 
 1 & 59.47 & 60.15 & 61.76 & 63.56\\ & 
 2 & 59.42 & 60.48 & 62.37 & 64.37\\ & 
 3 & 59.23 & 60.30 & 62.46 & 64.67  \\& 
 4 & 58.97 & 60.13 & 62.36 & 64.39 \\ 
 \bottomrule
\end{tabular} 
 \label{tab:k-hop}
\end{table}
\begin{figure*}[t]
    \centering
    \includegraphics[width=0.9\textwidth]{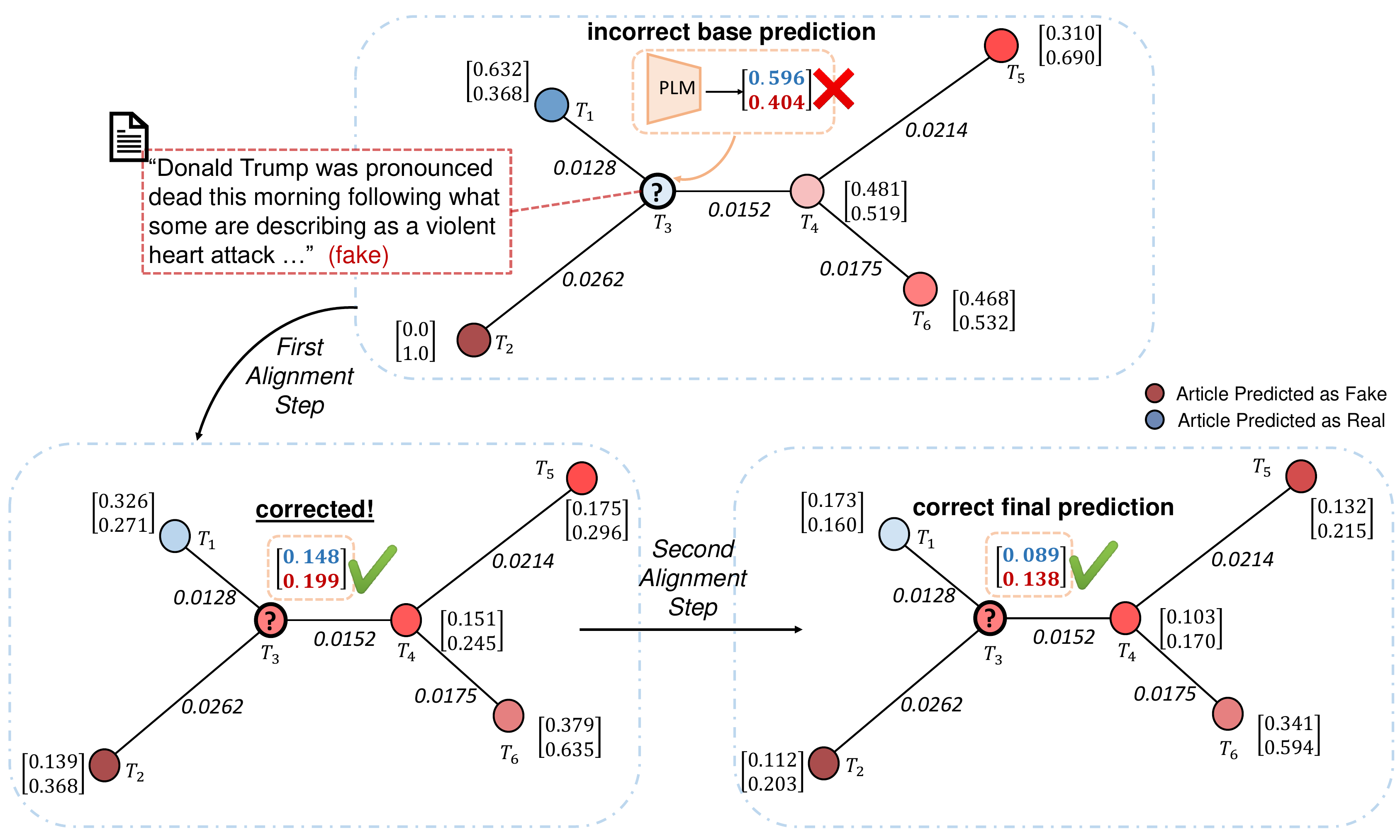}
    \caption{$\mathsf{P\&A}$ corrects the base predictions via confidence-aware prediction alignment on the news proximity graph. This can be observed via the logits assigned to the class labels throughout different alignment steps. The node with a question mark represents a news article to be classified, presenting an empirical illustration of how $\mathsf{P\&A}$ combines information from (1) prompt-based predictions, (2) confidence-informed label knowledge, and (3) social user engagements. For simplicity purposes, we only visualize the edges with highest weights for each node (the general illustration of $\mathsf{P\&A}$ is presented in Figure \ref{fig:framework}).}
    \label{fig:case-study}
\end{figure*}

\subsection{Impact of Prompt Design}
\label{sec:prompt_templates}
To evaluate the effectiveness of the prompt-based paradigm utilized in $\mathsf{P\&A}$, we explore the performance of $\mathsf{P\&A}$ across varying prompt templates (denoted as P1-3, respectively). For a fair evaluation, the mapping from prompting output to class labels is kept consistent across P1-3 as ``news''$\rightarrow0$ (real) and ``rumor''$\rightarrow1$ (fake). Note that we report the results of $\mathsf{P\&A}$ using P1 in all other subsections.

\begin{itemize} [leftmargin=*]
    \item \textbf{P1}: \underline{\textsf{[MASK]}}: <news article input> 
    \item \textbf{P2}: Contains \underline{\textsf{[MASK]}}: <news article input> 
    \item \textbf{P3}: Article with \underline{\textsf{[MASK]}}: <news article input> 
\end{itemize}

From Table \ref{tab:prompt-templates}, we observe that \textit{\textbf{(1)}} $\mathsf{P\&A}$ variants across multiple templates consistently outperform the most competitive baseline, which validates the effectiveness of prompting in eliciting pre-trained knowledge in PLMs for downstream tasks. \textit{\textbf{(2)}} With the increase of labeled samples, the fluctuation across different prompt templates is reduced, indicating that the need for a well-designed task-specific template is greatly eased with more annotated data.

\begin{table}
\caption{$\mathsf{P\&A}$ accuracy (\%) under different prompt templates. (B: best baseline acc. from Table \ref{tab:performance}; P1-3: varying prompts)
}
\centering 
 \begin{tabular}{lccccc} \toprule
 Dataset &  & $n=16$ & $n=32$ & $n=64$ & $n=128$\\
 \toprule
 \multirow{4}{*}{PolitiFact} & 
 B & 68.27 & 70.21 & 80.40 & 84.46  \\ &
 P1 & 85.30 & 83.73 & 86.79 & 89.45  \\ &
 P2 & 73.59 & 82.60 & 87.06 & 89.10  \\ &
 P3 & 78.57 & 81.81 & 86.98 & 89.66 \\
 \midrule
 \multirow{4}{*}{FANG} & 
 B & 56.87 & 56.46 & 60.36 & 61.85  \\ &
 P1 & 59.42 & 60.48 & 62.37 & 64.37\\ & 
 P2 & 59.01 & 58.96 & 63.43 & 64.33\\ & 
 P3 & 59.19 & 61.03 & 62.47 & 64.66  \\
 \bottomrule
\end{tabular} 
\label{tab:prompt-templates}
\end{table}

\subsection{Case Study}
\label{sec:case_study}

To further illustrate why our $\mathsf{P\&A}$ paradigm outperforms the most competitive baselines, specifically regarding $\mathsf{P\&A}$'s effectiveness of prediction alignment over the news proximity graph, we conduct a case study involving a test sample from the PolitiFact dataset (Figure \ref{fig:case-study}) that demonstrates $\mathsf{P\&A}$'s capability of correcting base prediction errors. 

Given an evidently fake news article that claims ``Donald Trump was pronounced dead this morning'' (denoted as $T_3$ in the Figure), the news content based ``Prompt'' component of $\mathsf{P\&A}$ mistakenly assigned it with a probability score of 0.596 for the ``real'' class, producing an incorrect base prediction. However, in the news proximity graph, $T_3$ is closely connected to multiple 1-hop and 2-hop neighbors ($T_2,T_4,T_5,T_6$) with predictions that favor the ``fake'' class at different confidence levels, in line with our hypothesis that similar readership implies similar news veracity. Among this neighborhood subgraph, the highest weighted edge connects the misclassified $T_3$ with a ground truth one-hot label at $T_2$. The edge weights of our news proximity graph are derived on the basis of the news articles' readership similarity; hence, $\mathsf{P\&A}$'s prediction alignment across these edges facilitates prediction consistency in terms of veracity. As we can see from the first alignment step that aggregates the prediction scores from each node's 1-hop neighbors, the high-confidence label knowledge from the ground-truth node $T_2$ effectively propagated to $T_3$, correcting $T_3$'s prediction scores to 0.148 for the ``real'' class and 0.199 for ``fake''. Besides this, the signals from other ``fake'' nodes ($T_4,T_5,T_6$) in the same subgraph are also reinforced via their respective 1-hop neighbors. Specifically, consider $T_4$, whose base prediction (i.e. class probabilities) only favors ``fake'' over ``real'' by a mere 0.02. Nevertheless, in the first alignment step, $T_4$ benefited from the soft label knowledge from higher-confidence neighbors $T_5$ and $T_6$, thereby gaining a more confident prediction of 0.151 for ``real'' and 0.245 for ``fake''. Similarly, in the second alignment step, the corrected ``fake'' prediction of $T_3$ remains unchanged, and $T_3$'s veracity-related signals are further enhanced, leading to improved results at the subsequent final classification stage.

Our case study shows that despite the abundant pre-trained knowledge in PLMs, base prediction errors can still arise due to the emergent nature of news topics. This highlights the necessity of supplementing the base predictions from PLMs with veracity-indicative evidence from the social context. 
Through aligning the prompting predictions on a veracity-consistent news proximity graph, $\mathsf{P\&A}$ effectively enhances fake news detection performance under label scarcity, and offers potential explainability in terms of users' news consumption preferences.

\section{Conclusion and Future Work}
In this paper, we investigate the fake news detection problem under a practical few-shot setting. We introduce \textit{``Prompt-and-Align''} ($\mathsf{P\&A}$), a novel prompt-based paradigm for few-shot fake news detection that jointly leverages pre-trained knowledge from PLMs and the veracity-indicative news readership patterns.
$\mathsf{P\&A}$ alleviates the label scarcity bottleneck by directly eliciting task-specific knowledge from a PLM via textual prompts, and encodes the veracity-consistent user engagement patterns with a news proximity graph. These innovations enable $\mathsf{P\&A}$ to fully exploit the limited high-quality labels and predictions, specifically via aligning the predictions over the news proximity graph in a confidence-informed manner. Extensive experiments on three real-world benchmark datasets validate the effectiveness of $\mathsf{P\&A}$ in terms of consistent performance gains across varying hyperparameter settings and prompts. Our work demonstrates promising potential for moving beyond the existing paradigms for fake news detection, 
suggesting more focused research on generalizable prompt-based learning frameworks and veracity-aware social context modeling schemes under the few-shot scenario.  

\section{Acknowledgements}
This research was supported by NUS-NCS Joint Laboratory (A-0008542-00-00). 

\bibliographystyle{ACM-Reference-Format}
\bibliography{reference}

\appendix

\end{document}